\documentclass[paper=a4]{scrartcl}

\usepackage[margin=1in]{geometry}
\usepackage[T1]{fontenc}
\usepackage[utf8]{inputenc}
\usepackage{lmodern}
\usepackage{microtype}

\usepackage{amsmath}
\usepackage{amssymb}
\usepackage{graphicx}
\usepackage[font=small]{caption}
\usepackage{mathtools}
\mathtoolsset{showonlyrefs}

\usepackage[authoryear,round]{natbib}
\usepackage{xcolor}

\usepackage[colorlinks,allcolors=blue!60!black]{hyperref}

\usepackage{authblk}

\newcommand{\boldf}{\mathbf{f}}

\newcommand{\bx}{\mathbf{x}}
\newcommand{\by}{\mathbf{y}}

\newcommand{\bs}{\mathbf{s}}

\newcommand{\bfphi}{\boldsymbol{\phi}}
\newcommand{\bfpsi}{\boldsymbol{\psi}}

\newcommand{\normal}{\mathcal{N}}
\newcommand{\order}{\mathcal{O}}

\DeclareMathOperator*{\ELBO}{ELBO}
\DeclareMathOperator*{\bbE}{\mathbb{E}}

\newcommand{\condset}[1]{\mathcal{C}_{#1}}
\newcommand{\anc}[1]{\mathcal{A}_{#1}}
\newcommand{\ranc}[1]{\mathcal{R}_{#1}}

\begin{document}

\title{Scalable Sparse Autoregressive Variational Inference for Non-Gaussian Spatial Fields\thanks{Short paper for the Joint Meeting SIS-FENStatS 2026, Rome, June 22-25, 2026.}}

\author[1]{Paul F.V. Wiemann}
\author[2]{Matthias Katzfuss}
\affil[1]{Department of Statistics, The Ohio State University, Columbus OH 43210, USA}
\affil[2]{Department of Statistics, University of Wisconsin--Madison, Madison WI 53706, USA}
\date{}

\maketitle

\begin{abstract}
\noindent
Latent field models are essential for spatial statistics, but the posterior distribution of the latent field, even with a Gaussian process (GP) prior, is analytically intractable whenever the likelihood is non-Gaussian.
Existing approximations
are either too slow or too restrictive for large non-Gaussian datasets.
We present a scalable variational inference framework that mirrors the sparse nearest-neighbor autoregressive structure of the prior in the variational family, uses normalizing-flow conditionals for expressiveness, and amortizes all variational parameters via a single shared neural conditioner.
Doubly stochastic optimization with reduced ancestor sets achieves polylogarithmic complexity per iteration.
\medskip

\end{abstract}

\section{Introduction}
Latent probabilistic fields appear throughout spatial statistics, for example
	as generalized GPs for classification and count data \citep{Diggle1998,Williams1998},
	in log-Gaussian Cox processes for spatial point patterns \citep{Diggle2013},
	modeling censored spatial observations \citep{Schelin2014spatialPrediction}, and
	in autoregressive GP-based models for non-Gaussian generative learning \citep{Katzfuss2021,Wiemann2023BayesianFields}.
In all of these settings, the posterior $p(\boldf\mid\by)$ of the latent field $\boldf$ given data $\by$ is analytically intractable, and scalable approximate inference is essential.

Standard approximations all have significant drawbacks at scale.
MCMC is exact in the limit but prohibitively slow for large $n$.
Laplace approximations impose a Gaussian posterior, severely underestimating non-Gaussian uncertainty \citep{nickisch2008approximationsBinaryGP}.
Mean-field VI \citep{Blei2017} ignores posterior dependence.
The sparse-inverse-Cholesky (SIC) Gaussian variational family of \citet{Cao2023} captures structured dependence but remains limited to Gaussian posteriors.

We address all of these limitations simultaneously.
Our framework
	(i) inherits the linear-cost structure of the Vecchia-type approximation,
	(ii) uses normalizing flows as conditionals for non-Gaussian expressiveness, and
	(iii) can amortize parameters via a shared neural conditioner, together enabling inference at scale.

\section{Sparse Autoregressive Framework}

\subsection{NN-AR prior}

Let $\boldf = (f_1, \dots, f_n)^\top$ denote a latent function evaluated at $n$ spatial
locations $\bx_1, \dots, \bx_n$ arranged in maximin order \citep{Guinness2016a}.
The prior factorizes as
\begin{equation}
  p(\boldf) = \prod_{i=1}^{n} p(f_i \mid \boldf_{j < i}) \approx \prod_{i=1}^{n} p(f_i \mid \boldf_{\condset{i}}),
  \label{eq:prior}
\end{equation}
where the approximation replaces conditioning on all previous entries $\boldf_{j<i}$ by conditioning only on the nearest-neighbor subset $\boldf_{\condset{i}} \subseteq \boldf_{j<i}$, with $\condset{i}$ defined as
\begin{equation}
  \condset{i} = \bigl\{j < i : \operatorname{dist}(\bx_i, \bx_j) < \rho\,\ell_i\bigr\},
  \label{eq:conditioning}
\end{equation}
with $\ell_i = \min_{j<i}\operatorname{dist}(\bx_i,\bx_j)$ and sparsity parameter $\rho \geq 1$.
Since $|\condset{i}| = \order(\rho^D)$ is constant in $n$, this Vecchia or nearest-neighbor autoregressive (NN-AR) structure reduces the prior evaluation and simulation to $\order(n)$ and, in the special case of a GP prior, each conditional in~\eqref{eq:prior} is Gaussian, yielding near-exact accuracy for elliptic covariance kernels \citep{Schafer2020}.

\subsection{NN-AR variational family and scalable ELBO}
\label{sec:vf}

We mirror the NN-AR prior structure in the variational family,
\begin{equation}
  q_{\bfphi}(\boldf) = \prod_{i=1}^{n} q_{\bfphi}(f_i \mid \boldf_{\condset{i}}),
  \label{eq:gen-var-fam}
\end{equation}
and optimize variational parameters $\bfphi$ by maximizing the evidence lower bound $\ELBO(\bfphi) = \bbE_{q_{\bfphi}}\!\bigl[\log p(\by\mid\boldf) + \log p(\boldf) - \log q_{\bfphi}(\boldf)\bigr]$.
The ELBO decomposes as a sum over $i=1,\dots,n$, where each term depends on $\boldf$
only through the ancestor set $\anc{i}$ of node $i$ in the NN-AR directed acyclic graph \citep{Katzfuss2017a}, so that $\ELBO(\bfphi) = \sum_{i=1}^{n} \mathbb{E}_{q_{\bfphi}}\left[g_i(\boldf_{\anc{i}}, \bfphi)\right]$ where $g_i$ collects the likelihood, prior, and variational density contributions of index $i$.

\begin{figure}[b]
  \centering
  \includegraphics[width=\linewidth]{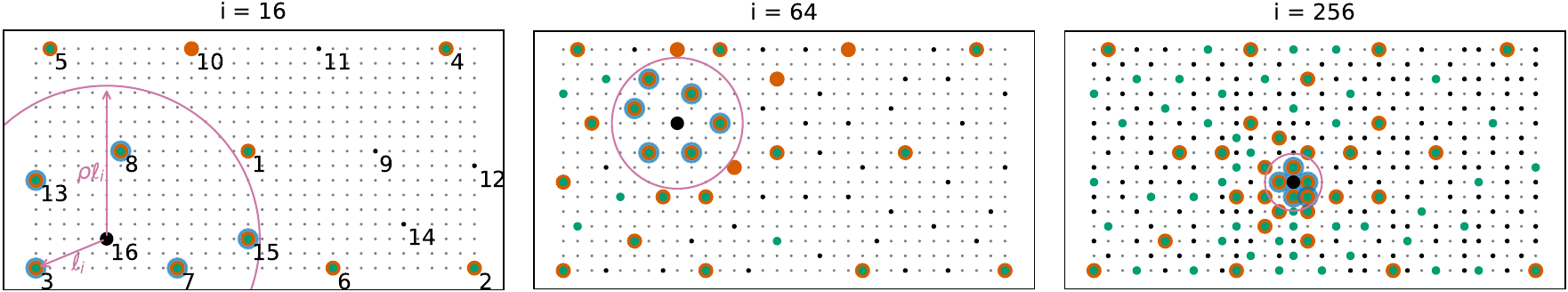}
  \caption{
  Maximin ordering of $n=512$ points. For selected indices $i$ (black dot), we show the conditioning set $\condset{i}$ (blue), full ancestor set $\anc{i}$ (green), and reduced ancestor set $\ranc{i}$ (orange). Clearly for large $i$, $|\anc{i}| > |\ranc{i}|$.
  }
  \label{fig:maxmin}
\end{figure}

\begin{figure}[t]
  \centering
  \includegraphics[width=0.4\linewidth,trim=0em 2em 0em 0em]{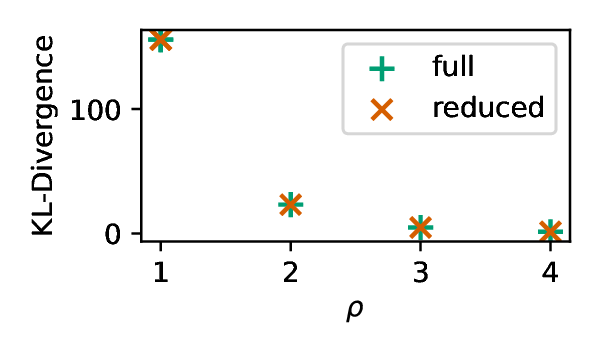}
  \caption{KL divergence to the exact posterior distribution vs.\ $\rho$ for noisy GP regression ($n=50$), using full and reduced ancestor sets. Accuracy improves monotonically with~$\rho$.}
  \label{fig:rho-kl}
 \end{figure}

\paragraph{Reduced ancestor sets.}
Naively sampling $\boldf_{\anc{i}}$ costs $\order(n)$ per step.
However, reduced ancestor sets $\ranc{i} = \{j<i : \operatorname{dist}(\bx_i,\bx_j) < \rho\,\ell_j\}$ have size ${|\ranc{i}|=\order(\rho^D\log n)}$ and suffice for accurate gradient estimation \citep{Cao2023}.
Moreover, the ELBO decomposition allows for using a mini-batch $\mathcal{B}$ of size $|\mathcal{B}|$ and $S$ Monte Carlo samples in a doubly stochastic gradient estimator
\begin{equation}
  \nabla_{\bfphi} \ELBO(\bfphi) \approx
    \frac{n}{|\mathcal{B}|}\sum_{i\in\mathcal{B}}
    \frac{1}{S}\sum_{s=1}^{S} \nabla_{\bfphi}g_i\!\bigl(\boldf^{(s,i)}_{\ranc{i}},\bfphi\bigr),
  \label{eq:dselbo}
\end{equation}
with gradients computed via automatic differentiation and the reparametrization trick
\citep{Kingma2014}. The resulting complexity per iteration is $\order(\rho^D\log n)$. %
As shown in Figure~\ref{fig:rho-kl}, accuracy improves monotonically with $\rho$.
Figure~\ref{fig:maxmin} illustrates maximin ordering, conditioning sets, and reduced
ancestor sets.

\section{Expressive Amortized Variational Family}

\subsection{Normalizing flow conditionals}

The framework in Section~\ref{sec:vf} is general. Any sampleable and differentiable conditional $q_{\bfphi}(f_i\mid\boldf_{\condset{i}})$ can be used.
We parameterize each conditional via a sparse autoregressive normalizing flow (ARNF),
\begin{equation}
  f_i = \tau(u_i,\bfpsi_i), \quad \bfpsi_i = \gamma_i(\boldf_{\condset{i}},\bfphi),
  \quad u_i \sim \normal(0,1),
  \label{eq:sparse-arnf}
\end{equation}
where $\tau(\cdot, \bfpsi_i)$ is a strictly monotone transformation function (e.g., deep sigmoidal flow \citep{pmlr-v80-huang18d}) and $\gamma_i$ is a (neural) conditioner mapping the context $\boldf_{\condset{i}}$ to flow parameters $\bfpsi_i$.
Under regularity conditions, ARNFs can be universal density approximators \citep{bogachev2005triangular}.
As a consequence, for sufficiently large $\rho$, the variational family can approximate any posterior arbitrarily well.
The family contains mean-field VI ($\rho=1$, $\tau$ affine) and the Gaussian SIC-VI of \citet{Cao2023} ($\tau(u_i) = m_i + d_i u_i$, with $\bfpsi_i = (m_i, d_i)$ where $m_i$ depends linearly on $f_{\condset{i}}$ and $d_i$ is constant w.r.t.\ $f_{\condset{i}}$) as special cases.

\subsection{Amortized conditioner}

Without amortization, each location $i$ has its own conditioner $\gamma_i$, giving
$\order(n)$ variational parameters in total.
We propose to share a single conditioner $\gamma_0$ across all locations, augmented with location-specific context $\bs_i$ (e.g., coordinates, neighbor distances, local observations), so
$
  \bfpsi_i = \gamma_0(\boldf_{\condset{i}}, \bs_i, \bfphi).
$

We use an encoder-pooling-decoder architecture, that is, an encoder maps each neighbor $(f_j, \bx_j)$ to a feature vector; an attention-pooling layer
aggregates features over $\condset{i}$; a decoder produces $\bfpsi_i$.
All variational parameters $\bfphi$ are updated in every mini-batch step.
This reduces the total parameter count from $\order(n)$ to $\order(1)$, enabling truly scalable inference.

\section{Results}

\begin{figure}[tb]
  \centering
  \includegraphics[width=\linewidth]{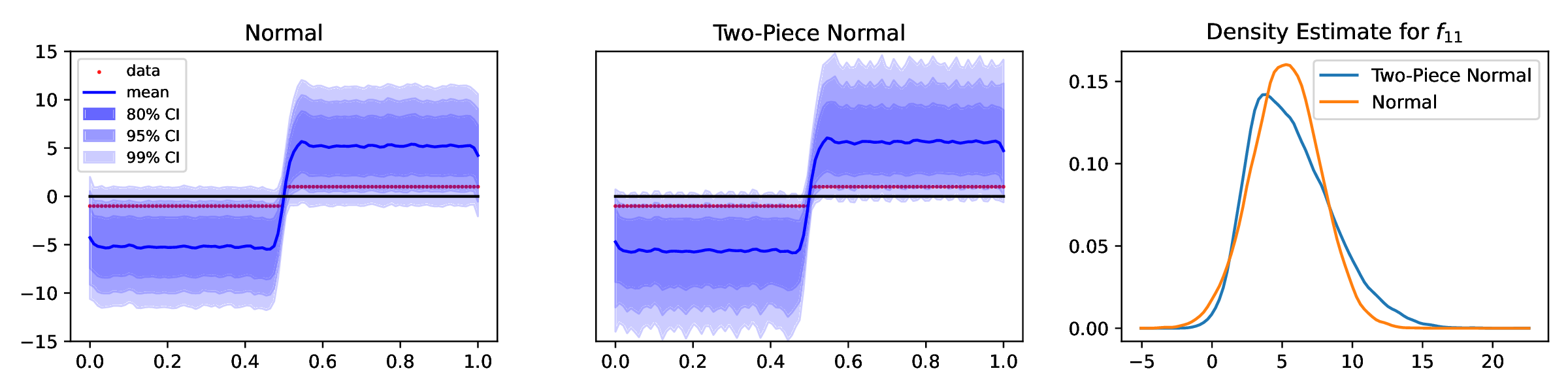}
  \caption{GP binary classification on the unit interval. Our two-piece-normal NN-AR family (middle) captures posterior skewness and achieves higher a ELBO than the Gaussian baseline (left).
  The right panel compares the estimated densities for $f_{11}$.
    }

  \label{fig:classification}
\end{figure}

\paragraph{GP classification.}
We applied the framework to binary data $y_i \in \{-1,+1\}$ with likelihood $P(y_i = 1 \mid f_i) = \Phi(f_i)$ and a GP prior. GP classification posteriors are well-known to be skewed \citep{nickisch2008approximationsBinaryGP}.
We compared a Gaussian SIC variational family against our family using two-piece-normal conditional distributions, which adds a per-location skewness parameter.
As shown in Figure~\ref{fig:classification}, our approach captures posterior skewness and achieves a higher ELBO.

\paragraph{Arctic Ocean ice coverage.}
We applied the amortized framework to binary Arctic Ocean ice-coverage data from NOAA's Interactive Multisensor Snow and Ice Mapping System, using a Mat\'ern-1.5 GP prior with probit likelihood.
Our amortized architecture with affine flow conditionals (as a first demonstration of amortization) requires only $1{,}398$ variational parameters regardless of $n$.

\begin{figure}[b]
  \centering
  \includegraphics[trim=0mm 10mm 0mm 4mm,clip,width=\linewidth]{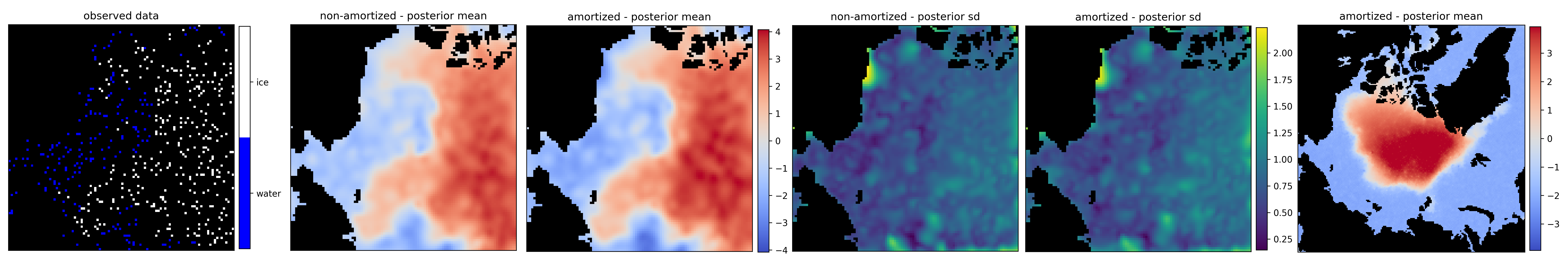}
  \caption{Left to right: Observed data, posterior mean (non-amortized and amortized), posterior standard deviation (non-amortized and amortized), posterior mean (larger dataset, amortized).
  }
  \label{fig:ice}
\end{figure}

For a subregion with $n=512$ randomly selected observations, the amortized approach visually matched the accuracy of a non-amortized version while using only $1{,}398$ vs.\ $4{,}580 \approx 9n$ parameters. 
Crucially, for a larger subset observed on a different date with $n=21{,}855$, the amortized approach produced accurate posterior maps of the latent field with only $1{,}398$ parameters, while the non-amortized version (approx.\ $200{,}000 \approx 9n$ parameters) did not converge reliably in our initial runs.
Figure~\ref{fig:ice} displays posterior summaries.

\section{Conclusion}
We have presented a scalable, expressive variational inference framework for non-Gaussian posterior distributions of latent spatial fields.
By combining sparse NN-AR structure, normalizing-flow conditionals, and neural amortization, the framework achieves polylogarithmic per-iteration complexity while reducing variational parameters to $\order(1)$ with amortized inference.
Preliminary results on GP classification and Arctic ice-coverage data demonstrate expressiveness and scalability of the approach.

\bibliographystyle{apalike}
\bibliography{abstract}

\end{document}